%% file: main.tex
\documentclass{article}
\usepackage[final]{nips_2017}

\usepackage{booktabs} % For formal tables
\usepackage{algorithm}
\usepackage{algorithmic}
\usepackage{dsfont}
\usepackage{amsfonts}
\usepackage{natbib}
\usepackage{url}
\usepackage[export]{adjustbox}
\usepackage{todonotes}
\usepackage{subcaption}

\newcommand{\task}{\mathcal{T}}

\newcommand{\loss}{\mathcal{L}}
\newcommand{\fitness}{\mathcal{F}}
\newcommand{\population}{\mathcal{P}}
\newcommand{\popupdate}{\mathcal{U}}
\newcommand{\getpopbatch}{\mathcal{S}}
\newcommand{\lossi}{\loss_{\task_i}}

\begin{document}
\title{Meta-Learning by the Baldwin Effect}

\author{Chrisantha Fernando, Jakub Sygnowski, Simon Osindero, Jane Wang, Tom Schaul, \\\textbf{Denis Teplyashin, Pablo Sprechmann, Alexander Pritzel, Andrei A. Rusu}\\
Google Deepmind,\\
London, UK\\
\texttt{chrisantha@google.com}
}

\maketitle

\begin{abstract}
The scope of the Baldwin effect was recently called into question by two papers that closely examined the seminal work of Hinton and Nowlan. To this date there has been no demonstration of its necessity in empirically challenging tasks. Here we show that the Baldwin effect is capable of evolving few-shot supervised and reinforcement learning mechanisms, by shaping the hyperparameters and the initial parameters of deep learning algorithms. Furthermore it can genetically accommodate strong learning biases on the same set of problems as a recent machine learning algorithm called MAML "Model Agnostic Meta-Learning" which uses second-order gradients instead of evolution to learn a set of reference parameters (initial weights) that can allow rapid adaptation to tasks sampled from a distribution. Whilst in simple cases MAML is more data efficient than the Baldwin effect, the Baldwin effect is more general in that it does not require gradients to be backpropagated to the reference parameters or hyperparameters, and permits effectively any number of gradient updates in the inner loop. The Baldwin effect learns strong learning dependent biases, rather than purely genetically accommodating fixed behaviours in a learning independent manner.

\end{abstract}

\input{baldwinian_meta}

\bibliographystyle{abbrv}
\bibliography{sample-bibliography}

\end{document}

%% file: baldwinian_meta.tex
\section{Introduction}

There is a growing interest in the machine learning community in meta-learning \cite{thrun1998learning}, i.e. learning to learn. Recently an influential model-agnostic meta-learning (MAML) algorithm was proposed for the fast adaptation of parameters in neural networks \cite{finn2017model}. It works by using gradient descent to learn the reference (initial) parameter values of a neural network from which new parameters can most rapidly be learned to solve a sample of tasks from a distribution of tasks. It requires a differentiable learning procedure to backpropagate into the reference parameter values, and even then it is limited in the number of gradient steps in the inner learning loop that can be made before second order gradient calculations become intractable. Meta-learning in this way achieves state of the art in few shot learning, for example by allowing a reinforcement learning algorithm to (within a few gradient updates) learn the optimal speed or direction of a simulated cheetah or four-legged robot based only on reward. 

Over a hundred years ago, a similar effect was proposed by Mark Baldwin \cite{baldwin1896new} to explain how evolution could deal with irreducibly complex adaptations without the need for Lamarckian information flow \cite{jablonka2014evolution}. John Maynard Smith described the effect as follows: ``If individuals vary genetically in their capacity to learn, or to adapt developmentally, then those most able to adapt will leave most descendants, and the genes responsible will increase in frequency. In a fixed environment, when the best thing to learn remains constant, this can lead to the genetic determination of a character that, in earlier generations, had to be acquired afresh in each generation'' \cite{jms1}.  In this formulation the Baldwin effect is really two effects, or a trade-off between two factors: initially genetically specified phenotypic plasticity (variance), followed by genetic accommodation of the induced trait (bias). Turney writes "...the Baldwin effect has two aspects. First, lifetime learning in individuals can, in some situations, accelerate evolution. Second, learning is expensive. Therefore, in relatively stable environments, there is a selective pressure for the evolution of instinctive behaviors.'' \cite{turney2002myths}. 

Here we compare these two algorithms -- MAML and the Baldwin effect -- on the same tasks. Note that unlike in MAML, our evolutionary experiments show no learning of the hyperparameters and initial parameters, only standard Darwinian evolution of these elements. We show that the Baldwin effect is competitive with MAML, biasing a learning algorithm to fit the distribution of tasks encountered during evolution, without some of the restrictions encountered in \cite{finn2017model} (e.g. having direct access to gradients). 

\begin{figure}
  \centering
\includegraphics[width=3.3in,keepaspectratio=True]{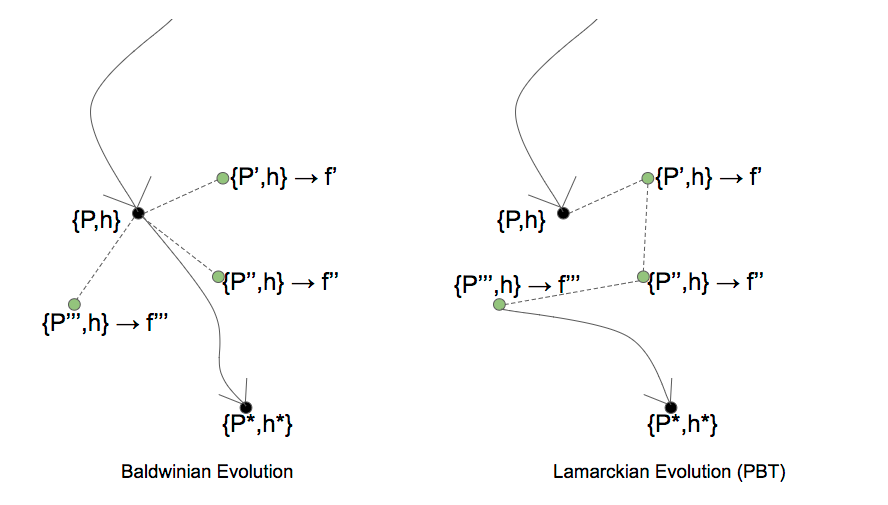}
\caption{Baldwinian evolution (left) versus Lamarckian evolution (right). In Baldwinian evolution the initial parameters $P$ and hyperparameters $h$ of a learning algorithm are evolved and subsequently evaluated by training on multiple independent learning trials (which adjust the weights $P$) to obtain the fitness. However, these learned parameters are not inherited by the next generation; only the original initial parameters $P$ and hyperparameters $h$ are inherited and then mutated to obtain $P^*$ and $h^*$. In contrast, in Lamarckian evolution the parameters are learned over multiple sequential learning trials and the final parameters $P'''$ (in the case of 3 learning trials shown in the diagram) are inherited along with the hyperparameters $h$, which are then mutated to produce $P^*$ and $h^*$. }
\label{fig:lamarckvbaldwin}
\end{figure}

In the framework of deep reinforcement learning (RL), evolution can be complementary to gradient descent by specifying and evolving the initial neural network parameters $P$ and hyperparameters $h$ of a learning algorithm \cite{castillo2006lamarckian, jaderberg2017population}. Throughout the course of learning, phenotypic plasticity is expressed as gradient updates are made to the parameters and the model learns to perform the task. This has the effect of smoothing the fitness landscape \cite{turney2002myths}. In the case of Baldwinian evolution, these updated weights are forgotten by the next generation, which instead inherit the initial weights $P$ and hyperparameters $h$, with possible mutation, whereas in Lamarckian evolution, these final, learned weights are evolved and passed on (see Figure \ref{fig:lamarckvbaldwin}). We refer to Darwinian evolution as the case where there is no learning within a lifetime. Lamarckianism closely resembles Population Based Training (PBT) \cite{jaderberg2017population}: a method for online hyperparameter evolution with the exception that in PBT we do not mutate the learned parameters $P$ but inherit them unchanged, only mutating hyperparameters: $h$ to $h*$. This method, while highly successful on a number of supervised, unsupervised and reinforcement learning tasks, has no incentive to learn a representation that can be easily evolved to solve a number of different tasks in a meta-learning setup. Control experiments in \cite{finn2017model} suggest that sequentially training (fine-tuning) a model on different tasks doesn't lead to competitive performance in meta-learning.

There is already evidence that the Baldwin effect has a role to play in machine learning because it is capable of evolving inductive bias in the form of the initial parameters $P$ and hyperparameters $h$ of a learning algorithm  \cite{castillo2006lamarckian, turney2002myths}. Here we test the more specific hypothesis that the Baldwin effect provides a way to evolve agents for few-shot data-efficient fast learning on a task distribution (previously examined mainly in classification settings \cite{ravi2016optimization, vinyals2016matching, santoro2016one}, but see also \cite{wang2016learning, duan2016RL2} for applications to RL), across a wide variety of learning domains. The Baldwin effect is thus used here as an algorithm for meta-learning, which results in a representation that is fit to a distribution of tasks, i.e. learning the structure of the various problems to be encountered rather than the specifics \cite{thrun1998learning}. The genome to be evolved is shaped by the task distribution, whereas the learning algorithm itself learns task specifics. The effect arises whenever there is a cost to learning imposed by the speed at which learning must occur. Such costs often arise in nature, for example in a co-evolutionary ecosystem where a newly born organism must rapidly learn to run so it can escape predators.

Our main experimental contributions are as follows: First we show that the Baldwin effect and MAML are comparable on a supervised learning task. Secondly we demonstrate that the Baldwin effect can be used in cases where MAML cannot be used, for instance in cases where the genotype is non-differentiable, e.g. where we evolve the macro-actions used by a discrete action RL algorithm, or the algorithms' discrete hyperparameters themselves. Thirdly we examine how genetic accommodation takes place in real deep neural networks undergoing the Baldwin effect. Fourthly, we examine two task distributions; one where Baldwinian learning is superior to Lamarckian learning and vice versa. 

\section{Related Work}

While deep learning systems trained with traditional supervised or reinforcement learning methods have achieved remarkable success in a variety of tasks, they perform poorly when only a small amount of data is available. Meta-learning aims to mitigate this limitation by broadening the learner's scope from a single task to a distribution of related tasks \cite{schmidhuber1996simple, thrun1998learning}. The goal of meta-learning is then to learn a learning strategy that generalizes to similar but unseen tasks from a given task distribution.
A lot of interest in meta-learning comes from the problem of one-shot learning in image classification, which
consists of learning a new class from a single labelled example \cite{lake2011one}.
Several approaches address this problem by using specialized neural network architectures that learn an embedding space that allows to effectively compare new examples. For instance, employing Siamese networks \cite{koch2015siamese} or recurrence with attention mechanisms \cite{vinyals2016matching}. These approaches achieve very good results in one-shot visual learning but cannot be easily employed in other tasks, such as reinforcement learning.
Another approach to meta-learning is to train a recurrent memory augmented learner to quickly adapt to new tasks of a given task distribution.
Such networks have been applied to few-shot image recognition \cite{santoro2016one} and reinforcement learning \cite{duan2016RL2,wang2016learning}.
More recent approaches propose to include the inductive bias of optimization-based learning into a meta-learner \cite{husken2000fast, ravi2016optimization, finn2017model}. 
Particularly related to this work is model-agnostic meta-learning (MAML) approach \cite{finn2017model}, that aims to learn the initial set of parameters such that they can be rapidly adapted (via gradient descent) to solve a given task from the task distribution. We describe this method in detail in section \ref{sec:algorithms}.

Hinton and Nowlan showed in their 1987 paper that the Baldwin effect works in the toy example of a needle-in-a-haystack binary optimization problem of 20 alleles (bits) using random search, where the random search distribution is encoded by evolution \cite{hinton1987learning}. The emphasis in their paper was to show how learning can smooth a (single task) rugged fitness landscape. The generality of that work has recently been called into question by a paper which claims that the scope of the effect is severely limited \cite{santos2015phenotypic}. Specifically, Santos et al showed that under certain task parameter settings (i.e. initial ratios of correct alleles $P(1) = 0.5$, and incorrect alleles $P(0) = 0.5$) then standard Darwinian evolution finds the needle in a haystack in the same number of generations as the Baldwin effect on average. However, a more recent paper by the same authors has shown that, when actually using Hinton and Nowlan's original conditions, $P(1) = 0.25$, $P(0)= 0.25$, and $P(?) = 0.5$, (where ? refers to an allele that does random search) the Baldwin effect is indeed significantly faster than Darwinian evolution \cite{fontanari2017revival}. In short, we know the Baldwin effect is possible in this toy task, only in a subset of the parameter conditions, but we wish here to provide convincing empirical support for the necessity of the Baldwin effect in more substantial and contemporary learning tasks than $L = 20$ bit learning problems using random search. 

The Baldwin effect in neural network learning has been investigated in several papers following the original work of Hinton and Nowlan \cite{downing2010baldwin}. Notably, Keesing and Stork used a genetic algorithm to evolve the initial weights of a neural network for digit classification, and found that the extent of genetic accommodation by the Baldwin effect depended on the amount of learning; too much learning and evolution was slowed down because there was too little selection pressure on the initial weights (as learning can do well form any starting position); too little learning and evolution was slowed down because fitness landscapes were not sufficiently smoothed \cite{keesing1991evolution}. Interestingly, they found that randomly sampling the number of gradient update steps from a distribution rather than using a fixed number significantly increased the rate of accommodation because that way the cost of learning was always felt by selection. Bullinaria evolved learning rate schedules during a lifetime, showing the evolution of developmental critical periods tailored to specific problems \cite{bullinaria2002evolution}. Neural network learning is only one kind of phenotypic plasticity. Turney used a genetic algorithm to evolve a population of biases for a decision tree induction algorithm for classification \cite{turney1995cost}. Cecconi et al evolved a hyperparameter determining how much learning, by imitation of a parent, an offspring will do in a co-evolutionary system \cite{cecconi1995maturation}. Anderson modelled how the adaptive immune system could facilitate natural antibody production by the Baldwin effect \cite{anderson1996adaptive}. Bull argued that the Haploid-Diploid cycle was a primitive example of learning and so subject to the Baldwin effect. \cite{DBLP:journals/corr/Bull16a}. This paper extends the existing literature by applying the Baldwin effect to deep learning reinforcement learning algorithms in task distributions.

\section{Methods}

% First the three algorithms that we compare are described. Secondly the models that we train and evolve are outlined, and finally we describe the three tasks we solve.   
In this section, we begin by describing the three algorithm families that we compare. Secondly we describe the three tasks that we solve, before finally outlining the details of the models that we train and evolve.

\subsection{Algorithms} 
\label{sec:algorithms}

{\bf Model-Agnostic Meta-Learning.} Given a distribution over tasks $p(\task)$ and a neural network with parameters collectively denoted $\theta$, MAML aims to learn an set of reference parameters $\theta^\ast$ such that one or a small number of gradient decent steps computed using a small amount of data for a given task from the distribution, $\task_i \sim p(\task)$, leads to effective generalization on that task.

The objective function for MAML is given by
\begin{equation}
\min_\theta \,\mathbb{E}_{\task_i \sim p(\task)}\, \lossi( \theta'_i) \, \\
\label{eq:maml}
\end{equation}
where the expectation is taken over the task distribution, $\lossi$ represents the loss corresponding to task $\task_i$ and the parameters $\theta'_i$ are the parameters adapted to fit $K$ representative training examples for this task. The task-specific learning is obtained via gradient descent,
\begin{equation}
\theta'_i = \theta-\alpha \nabla_\theta  \lossi(  \theta )
\nonumber
\end{equation}
where $\alpha$ is the learning rate. We used a single gradient descent step for ease of notation, but multiple steps can be used insted. The outer loss in (\ref{eq:maml}) evaluates the generalization of $\theta'_i$ on a small amount of validation data for the $i-$th task. The reference set of parameters $\theta^\ast$ are found by minimizing (\ref{eq:maml}) via stochastic gradient descent. The procedure is given in Algorithm~\ref{alg:maml}. Note that high order gradients are required to compute the parameter update.\\

\noindent {\bf Genetic Algorithm} (GA) is a general-purpose optimization algorithm inspired by the biological processes of mutation and selection. In our work, we use two flavors of Genetic Algorithms: a Steady State Genetic Algorithm and Generational Genetic Algorithm \cite{goldberg1991comparative}. In section \ref{sec:tasks} we introduce a sinusoid fitting task and a physics task domain. In the sinusoid fitting experiments we use a generational GA of population size 100, and rank-based selection. The physics-based RL experiments use an asynchronous parallel steady state GA with population size 500, and tournament size 10. The Baldwinian evolution algorithm hybridizes the GA and gradient-based learning as shown in Algorithm 2. We compare the Baldwinian algorithm with to two baselines: standard Darwinian evolution, and Lamarckian evolution.

\input{algos}

With some small modifications to Algorithm \ref{alg:baldwin_maml} we can obtain a learning process for single task or for a continual/multi-task learning setting by allowing all gradient updates to act successively on the same set of parameters. In this setting we can also consider a further modification/variant in which we use Lamarckian inheritance (changing line 15 to update the population using the parameters from the end of the short run of gradient optimization).\\

\noindent {\bf Natural Evolution Strategies} (NES) are a family of continuous black-box optimization algorithms that maintain and adapt a (Gaussian) search distribution in order to maximize the expected fitness under that distribution \cite{wierstra2008nes,wierstra2014jmlr};
they update the distribution parameters $\langle \mu, \Sigma \rangle$ in the direction of the (natural) policy gradient, as estimated from the fitnesses $f^g$ of a population of samples $\theta^{g} \sim \mathcal{N}(\mu, \Sigma)$. In our case specifically, we employ the variant called separable NES (SNES~\cite{Schaul2011snes}) that models only the element-wise variances $\sigma^2$, instead of the full covariance matrix $\Sigma$, because its linear complexity enables it to scale the high-dimensional spaces required for our experiments.

\subsection{Tasks}
\label{sec:tasks}

A supervised regression problem distribution and two physics-based reinforcement learning task distributions are used to compare the algorithms.\\

\noindent {\bf Sinusoid-fitting task:}
In this supervised task, in any one lifetime the agent must fit by regression a single sinusoid drawn from a distribution of phases and amplitudes. If evolution were to encode a non-plastic neural network, it would be impossible for it to do more than evolve the mean sinusoid for the distribution; whereas with lifetime learning, the initial function at birth could be modified to fit the sampled sinusoid encountered in any particular lifetime. In this case, the Baldwin effect would be expected to take place. This leads to a different perspective on the Baldwin effect to that taken by Hinton and Nowlan and others; whereby its role is not primarily for smoothing fitness landscapes in otherwise unsolvable adaptive problems, but instead for meta-learning distributions of adaptive problems that would be entirely unsolvable by evolution alone without phenotypic plasticity, and then encoding these distributions genetically to produce faster learning.

We compare the performance of MAML, NES, and generational GA on fitting sinusoids. In each generation, we sample $25$ different sine waves, out of which we select $10$ points for training and $10$ points for testing. The amplitude of sine waves was sampled uniformly from $[0.1, 5.0]$ and the phase from $[0, \pi]$. In one fitness evaluation, we perform $5$ gradient descent steps for each sine wave using training points, evaluate performance as mean-squared error on test data, and average the results for different sine waves. For NES and the GA, the fitness is the final MSE for that task after the gradient updates obtained over a separate sample of data than what was trained on. In both our models (generational Genetic Algorithms and Natural Evolution Strategies) the different genotypes in a given generation are evaluated using the same data, so that the amount of data our models see after some fixed number of generations is equal to the data baseline MAML sees after doing that number of meta-updates. In NES the population size was $25$ and in GA it was $100$.\\

\noindent {\bf Physics simulation reinforcement learning tasks.}
In reinforcement learning, the goal of few-shot meta-learning is to enable an agent to quickly acquire a policy for a new task based on training on the same distribution of tasks.  For example, an agent might learn to quickly run at a certain desired target speed or direction. We constructed two sets of tasks based on those used in Finn et al \cite{finn2017model}. One fitness evaluation consisted of 10 independent episodes with different task parameters. For the Baldwinian training condition, the parameters were reset to the inherited parameters at the start of each episode, and before inheritance. With Lamarckian training, the parameters were not reset between episodes and were inherited at the end of the final episode. In the Darwinian case there was no learning (gradient updates) in any of the 10 episodes. Fitness was defined as the sum of rewards obtained over all 10 episodes, providing an implicit selection pressure to learn quickly.

Two types of high-dimensional locomotion tasks were investigated using the MuJoCo simulator \cite{toderov}, specifically using the Planar Cheetah task, requiring it to run in either a particular direction (Goal Direction) or at a particular velocity (Goal Velocity). In the goal velocity experiments, the reward was the negative absolute value between the current velocity of the agent and a target velocity. The target velocity for each episode was chosen exhaustively in steps of 0.2, in the range 0.0 to 2.0. There were 10 such episodes for one fitness evaluation. The fitness was the sum of the rewards over these 10 episodes. In the goal direction experiments, the reward was the magnitude of the velocity in either the forward or backward direction. The fitness was the summed reward over 10 episodes with each episode alternating in whether backwards or forwards movement was required. In both cases the length of one episode was 3000 time steps (30 seconds) with a rollout size of 40 simulation time steps per gradient step, unless otherwise noted.

\subsection{Models}
\label{sec:models}
\noindent {\bf Sinusoid regression network:} The model architecture we used for the sinusoid fitting task (see Tasks) was the same as in \cite{finn2017model}: a neural network with two hidden layers with $40$ neurons each. We used Gaussian noise with mean $0$ and std $0.01$ to initialize the weights and biases of the network in GA and NES. The mean squared error loss is used to train the parameters of the network using stochastic gradient descent.\\

\noindent {\bf A2C Controller:} For the RL tasks described in section \ref{sec:tasks} we use the a2c algorithm (policy gradient with a trainable baseline) \cite{mnih2016asynchronous} to estimate the gradient for training the controller of the running cheetah agent. It consists of a shared torso which is a feed-forward neural network with rectified-linear transfer functions and two hidden layers of size 100. A policy readout from this final layer outputs a softmax over 12 possible discrete actions. A value function readout from the final layer outputs a single scalar value which is used as a baseline for the policy gradient algorithm. On replication, Gaussian noise of mean zero and standard deviation 0.02 is added to each weight and bias of the neural network. Additional noise is added to the hyperparameters which are the learning rate, entropy lost scale, and discount of the a2c algorithm.

For the physics-based RL tasks (see Tasks), macro-actions are evolved by the Baldwin effect; i.e. the 12x7 action primitive matrix which determines the 7 motor torques produced for each of the 12 discrete actions the cheetah controller can execute at each time step. The use of second order gradients to modify such hyperparameters is known to be notoriously unstable, whilst the Baldwin effect allows meta-optimization of these hyperparameters as well as the initial weights as in \cite{finn2017model}. In the a2c experiments we additionally explored the use of a genetically encoded binary vector of hyperparameters of the same length as the number of parameters in the model. This vector (which we call a mask) determines whether or not each parameter should be learnable or not. Bit flip mutation is used to evolve this mask hyperparameter vector. Thus is done in order to emulate the setup in the original Hinton and Nowlan paper.

\section{Results}

Firstly, MAML is compared with genetic algorithms and natural evolution strategies on a supervised learning task; fitting sinusoids drawn from a distribution. Secondly,  genetic algorithms are used to evolve the hyperparameters and initial parameters of a policy gradient algorithm with an adaptive critic, on two reinforcement learning problems.

\subsection{Rapid fitting of sinusoids}
\begin{figure}[h]
  \begin{subfigure}{0.5\textwidth}
\includegraphics[width=2.5in,keepaspectratio=True]{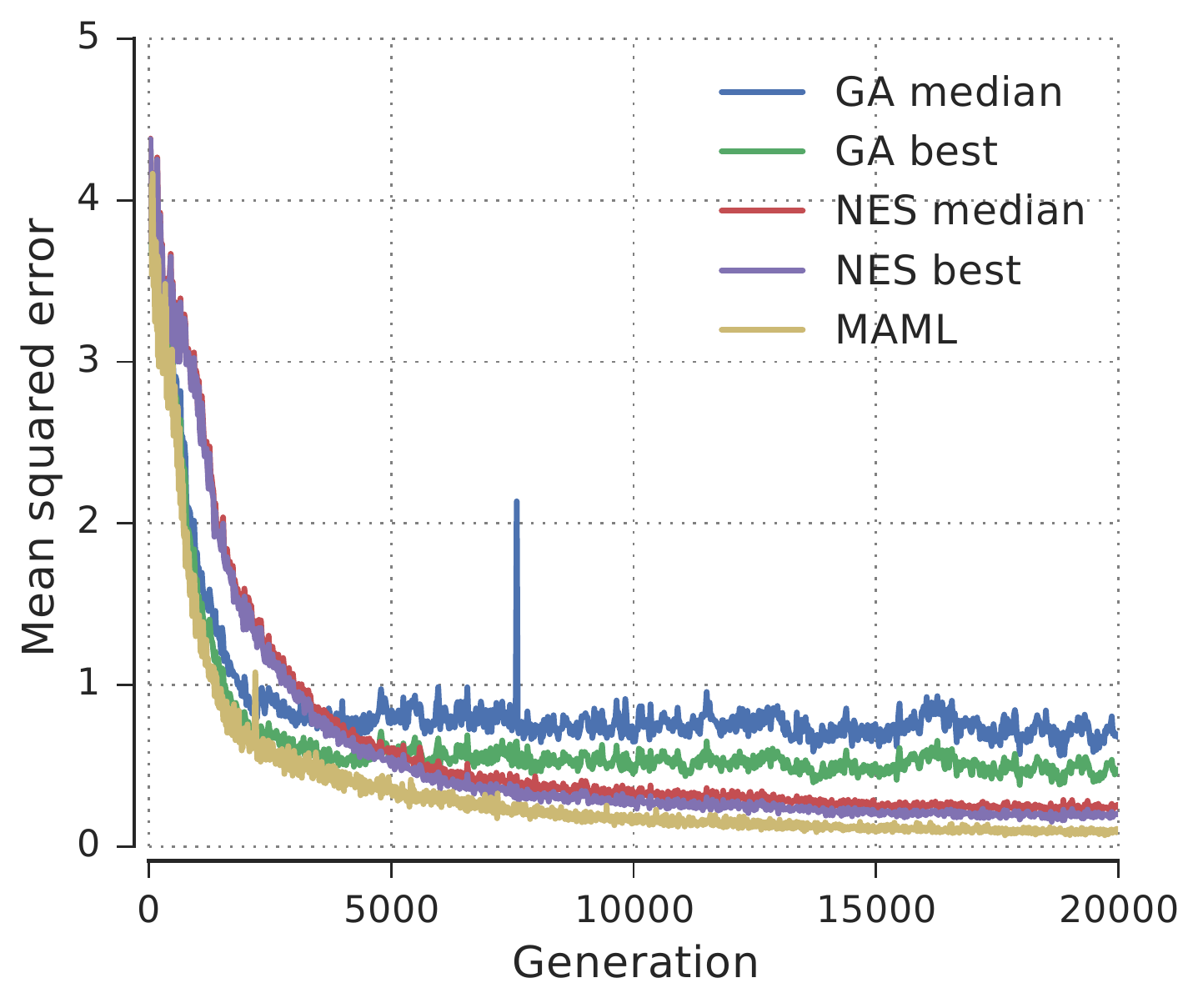}
\caption{Sinusoid fitting task learning curves show a comparison of mean-squared errors of the Baldwin effect operating over evolution in NES and GA, vs during MAML training. Despite using only final fitness as a training signal, our methods achieve results on par to MAML.}
\label{sine-ours}
\end{subfigure}
  \hspace{5mm}
  \begin{subfigure}{0.5\textwidth}
\includegraphics[width=2.5in,keepaspectratio=True]{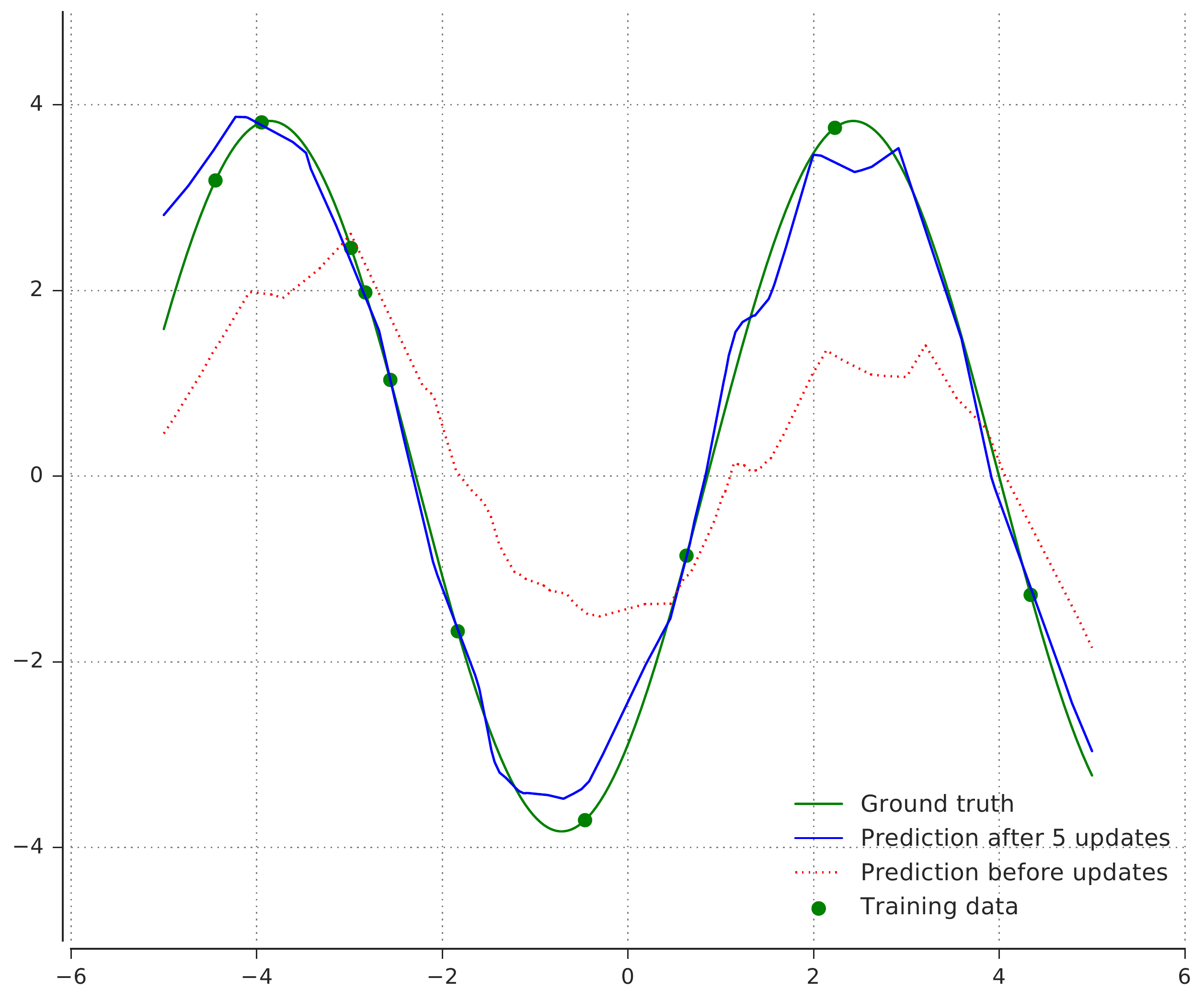}
\caption{Example fitting of a sine wave using trained NES model. The parameters correspond to the mean of the distribution after $20000$ generations. The red curve shows the initial function of the regression network prior to learning -- note that it has evolved to be a sine wave. The blue line shows the function learned after five gradient steps, and the green curve shows the target sine wave to fit in this particular lifetime.}
\label{sine-nes}
\end{subfigure}
  \caption{Results of training on the sine wave task.}
\end{figure}

The performance of MAML, NES and GA on fitting sinusoids is shown in Figure \ref{sine-ours}.  As our methods are based on a population of genotypes, we plot both the median and the best fitness achieved in each generation.

\begin{figure}[ht]
    \centering
\includegraphics[width=3.3in,keepaspectratio=True]{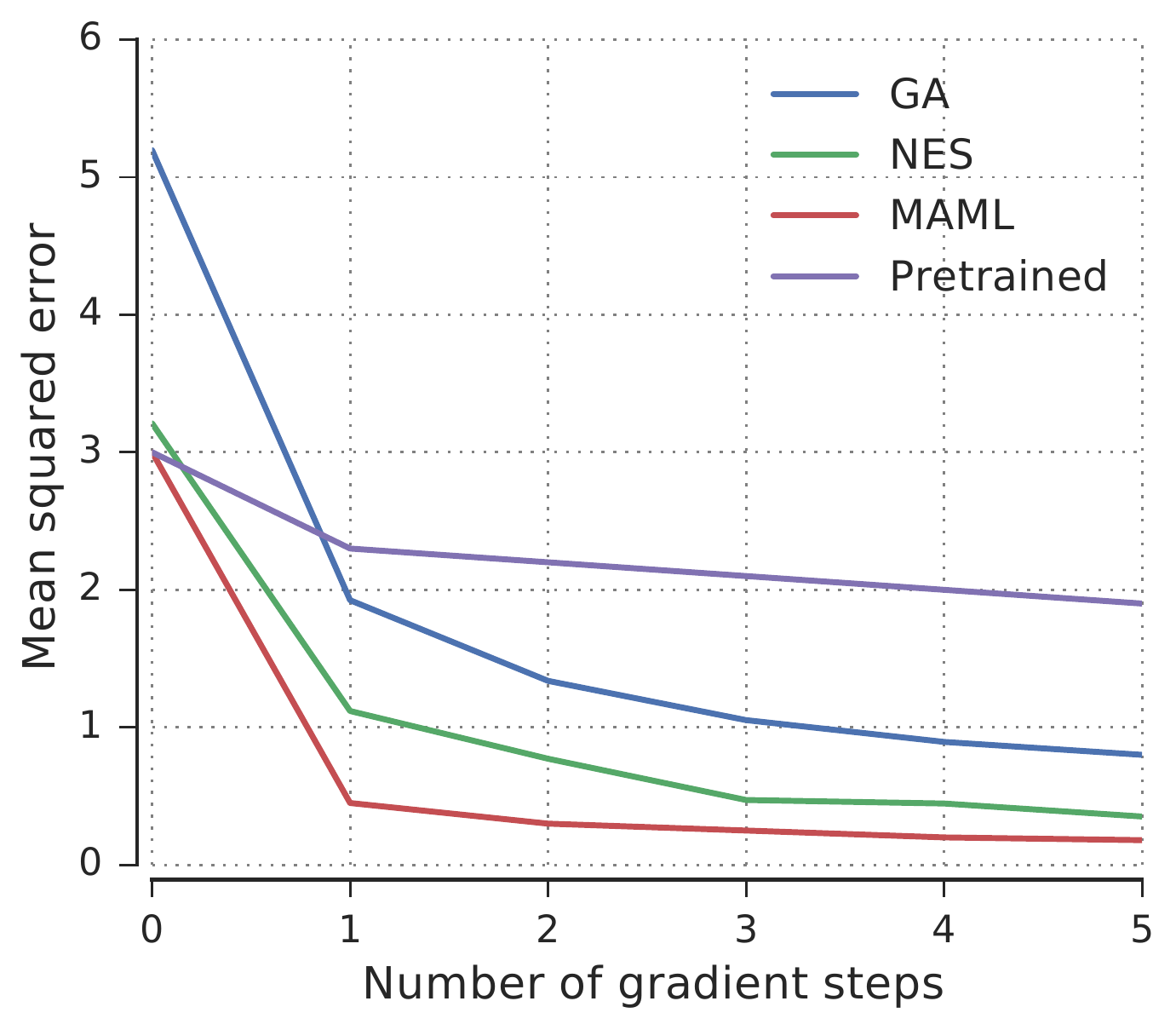}
\caption{Comparison of the speed of fitting of the sine wave during validation time, with MAML, NES, and GA.}
\label{sine-accomodation}
\end{figure}

\begin{figure}[ht]
  \begin{subfigure}{0.5\textwidth}
\includegraphics[width=.9\textwidth,keepaspectratio=True]{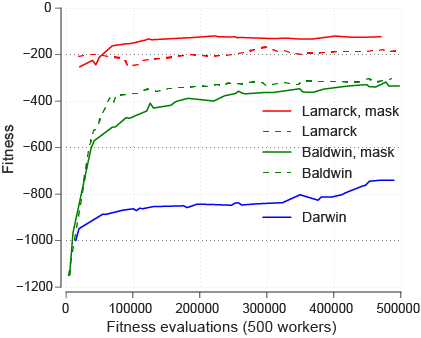}
\caption{Baldwin (Green), Lamarck (Red), Darwin (Blue) fitness on the Cheetah Goal Velocity task, showing that Lamarckian evolution works best when evaluated on tasks very similar to the previously learned on.}
\label{GoalVelocity}
\end{subfigure}
  \hspace{5mm}
  \begin{subfigure}{0.5\textwidth}
\includegraphics[width=.9\textwidth,keepaspectratio=True]{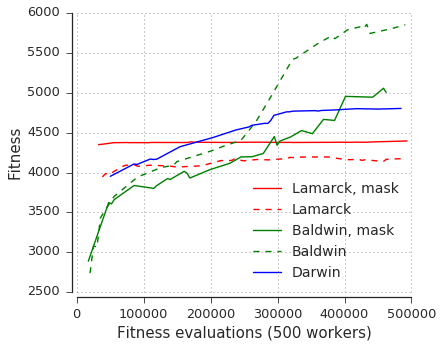}
\caption{Baldwin (Green), Lamarck (Red), Darwin (Blue) fitness on the Cheetah Goal Direction task.}
\label{GoalDirectionSummary}
  \vspace{5mm}
\end{subfigure}
  \caption{Results of training on the Cheetah tasks.}
\end{figure}

\begin{figure*}[h]
    \centering
  \begin{subfigure}{0.19\textwidth}
%    \subfloat[\scriptsize{Baldwin, mask}]{
    \includegraphics[width=0.9\textwidth,valign=c]{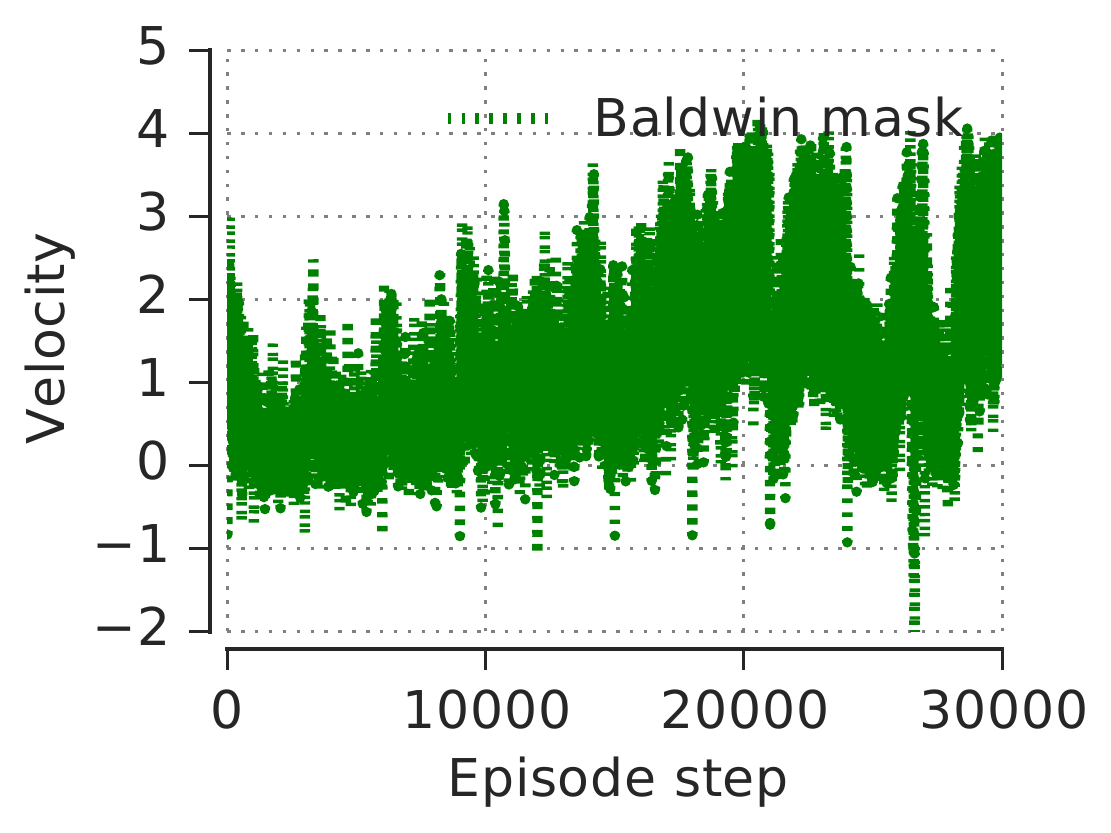}
    \caption{Baldwin, mask}
  \end{subfigure}
  \begin{subfigure}{0.20\textwidth}
%    }
%    \subfloat[\scriptsize{Baldwin, no mask}]{
    \includegraphics[width=0.9\textwidth,valign=c]{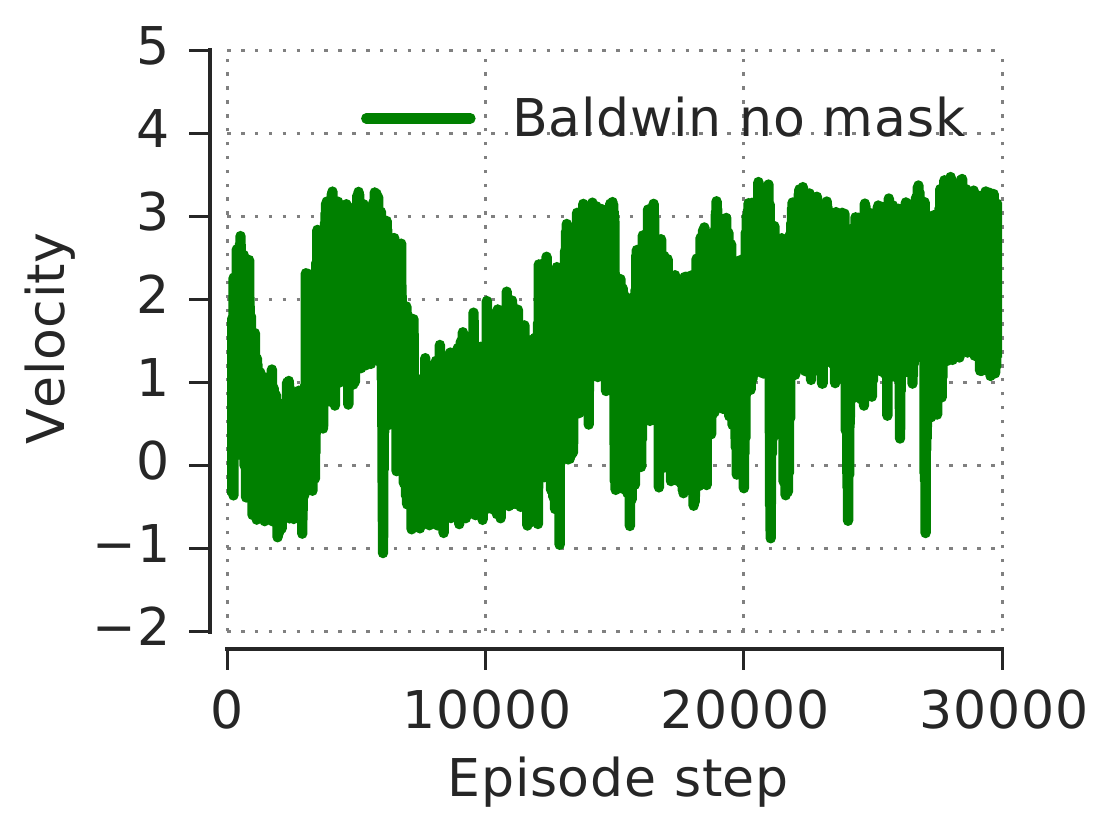}
    \caption{Baldwin, no mask}
  \end{subfigure}
  \begin{subfigure}{0.19\textwidth}
%    }
%    \subfloat[\scriptsize{Lamarck, mask}]{
    \includegraphics[width=0.9\textwidth,valign=c]{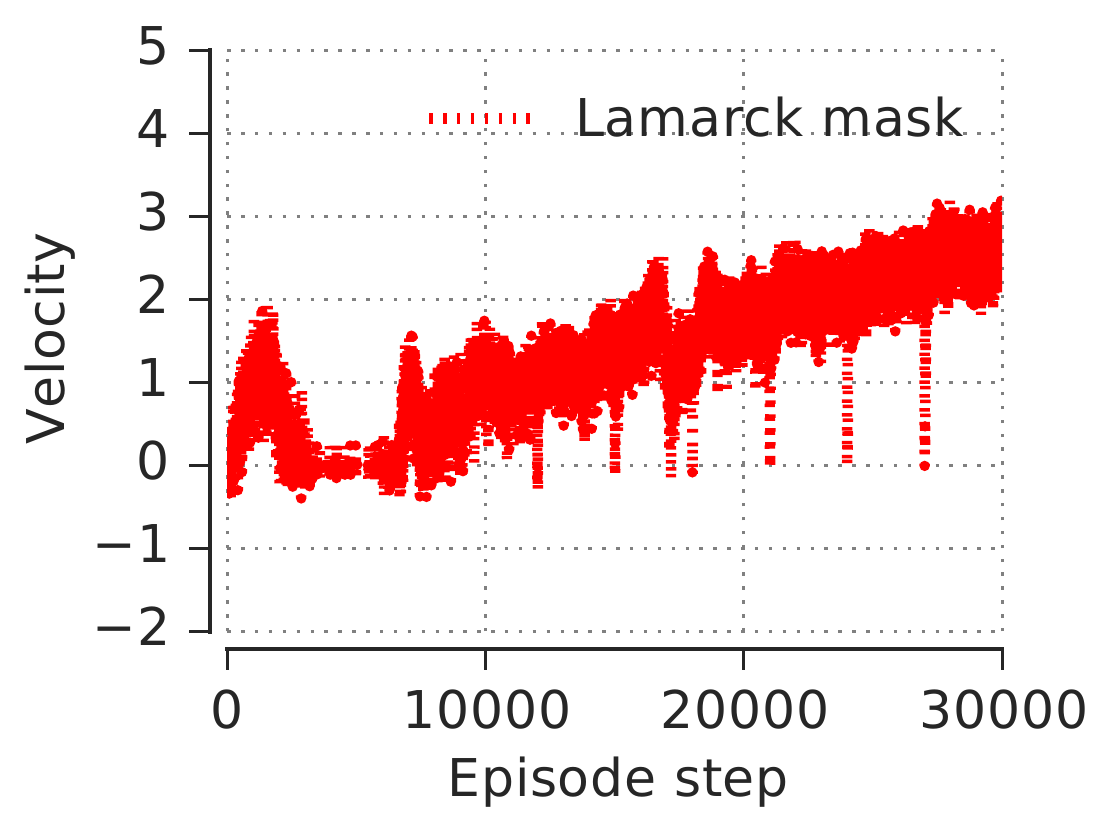}
    \caption{Lamarck, mask}
  \end{subfigure}
  \begin{subfigure}{0.20\textwidth}
%    }
%    \subfloat[\scriptsize{Lamarck, no mask}]{
    \includegraphics[width=0.9\textwidth,valign=c]{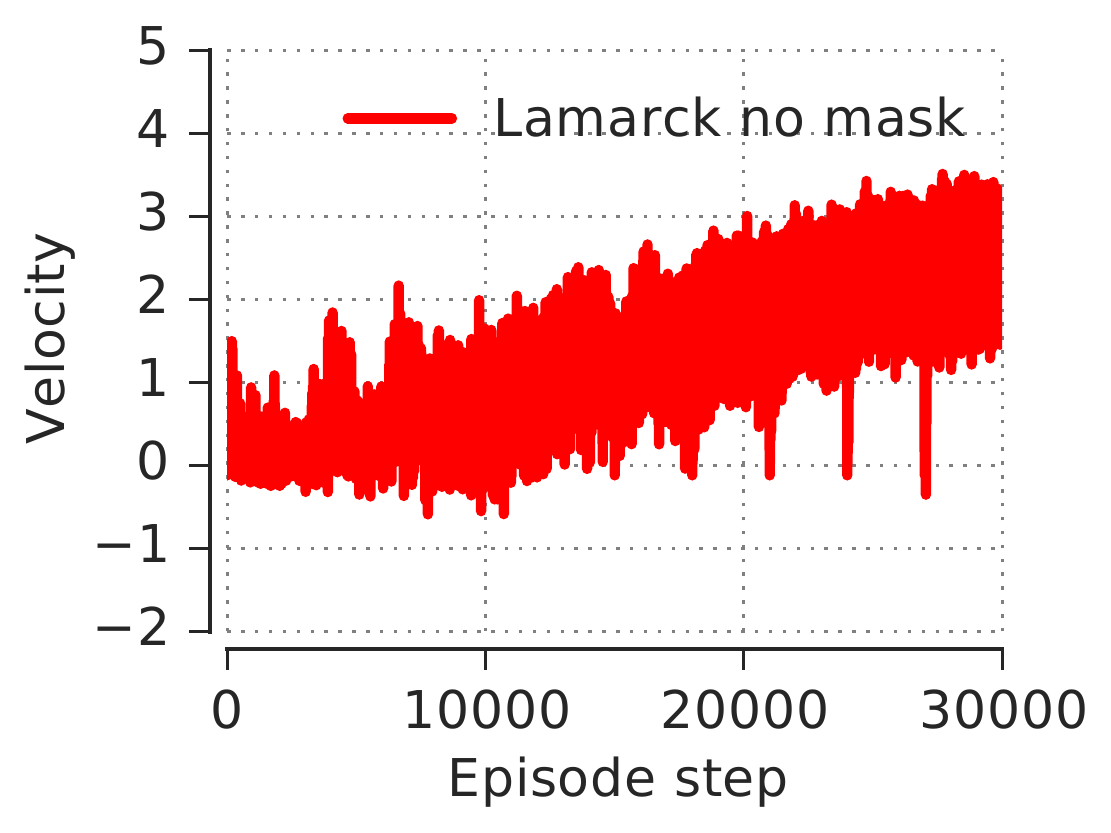}
    \caption{Lamarck, no mask}
  \end{subfigure}
  \begin{subfigure}{0.19\textwidth}
%    }
%    \subfloat[\scriptsize{Darwin}]{
    \includegraphics[width=0.9\textwidth,valign=c]{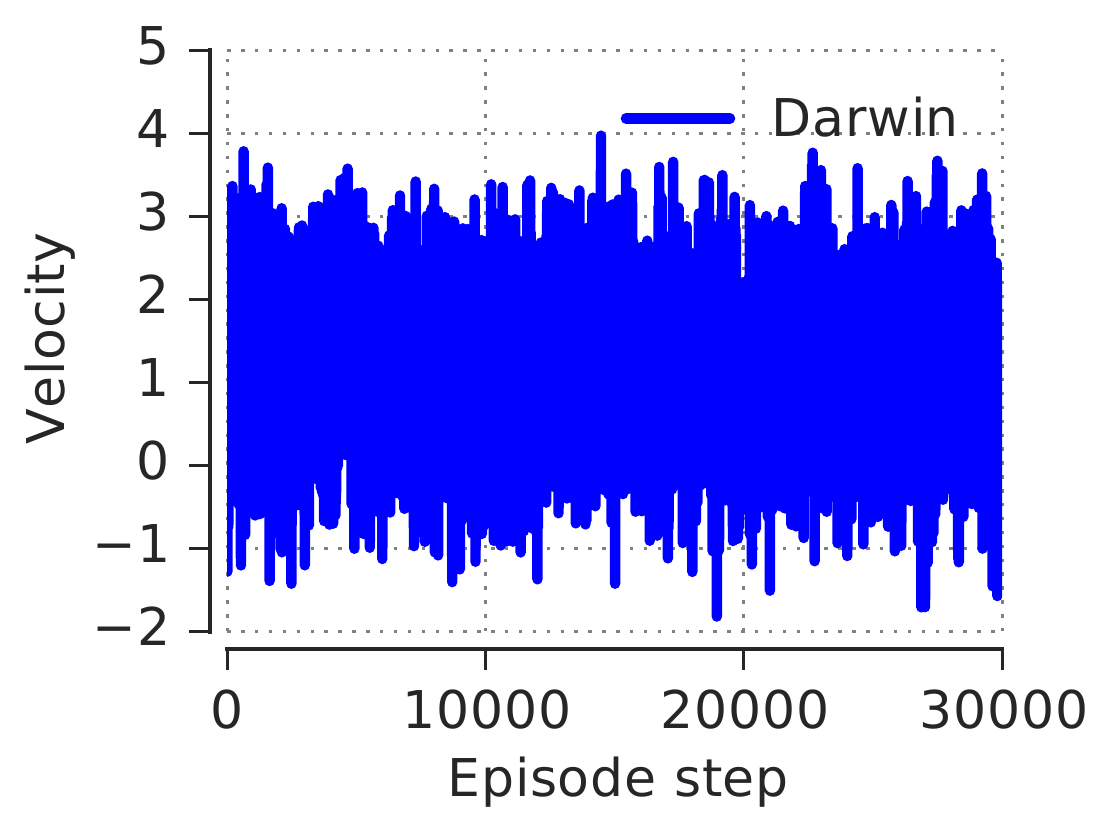}
    \caption{Darwin}
  \end{subfigure}
%    }
    \caption{Velocities obtained in the Cheetah Goal Velocity task for (a,b) Baldwinian, (c,d) Lamarckian, and (e) Darwinian evolution. Baldwinian evolution tends to revert back to low velocities during training, while Lamarckian maintains a steadily increasing forward velocity, as a result of sequence of tasks trained on (target velocity incremented by 0.2 every episode). Video1Supp shows the Lamarckian agent running forwards at different speeds.}
    \label{GoalVelocityProfile}
\end{figure*}

\begin{figure*}[ht]
    \centering
  \begin{subfigure}{0.19\textwidth}
    \includegraphics[width=0.9\textwidth,valign=c]{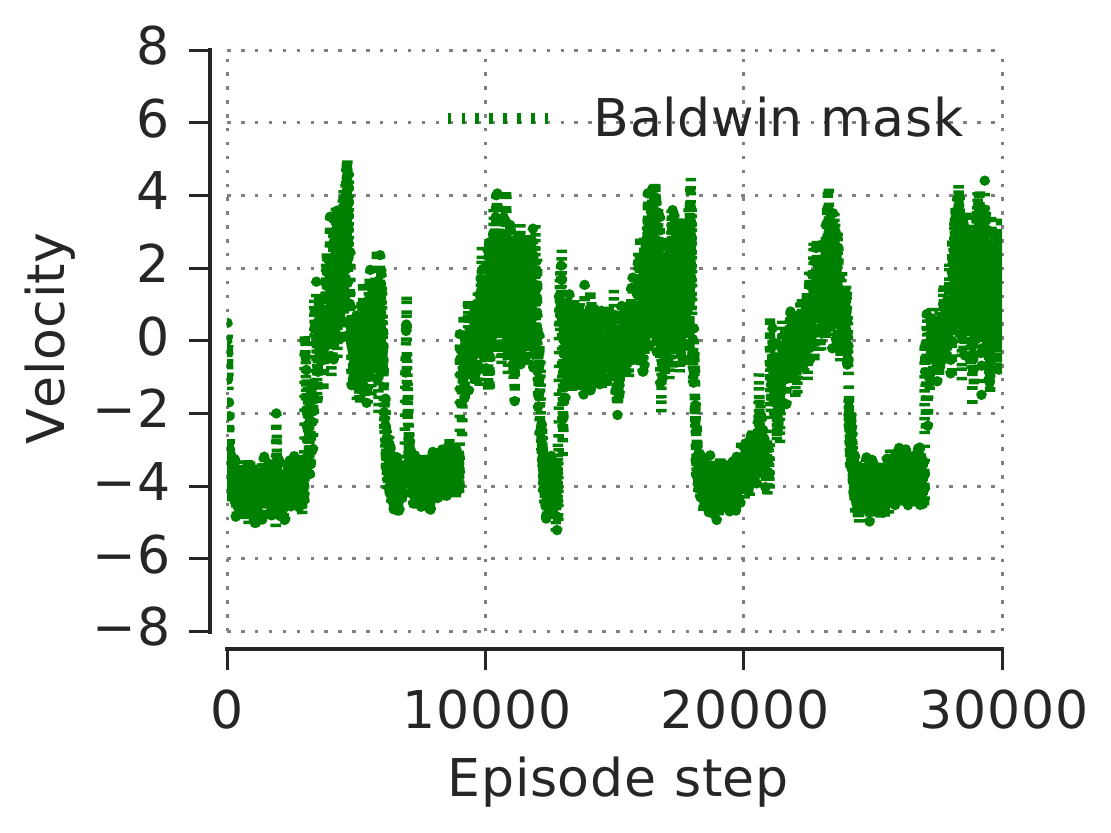}
    \caption{Baldwin, mask}
  \end{subfigure}
  \begin{subfigure}{0.20\textwidth}
%    }
    \includegraphics[width=0.9\textwidth,valign=c]{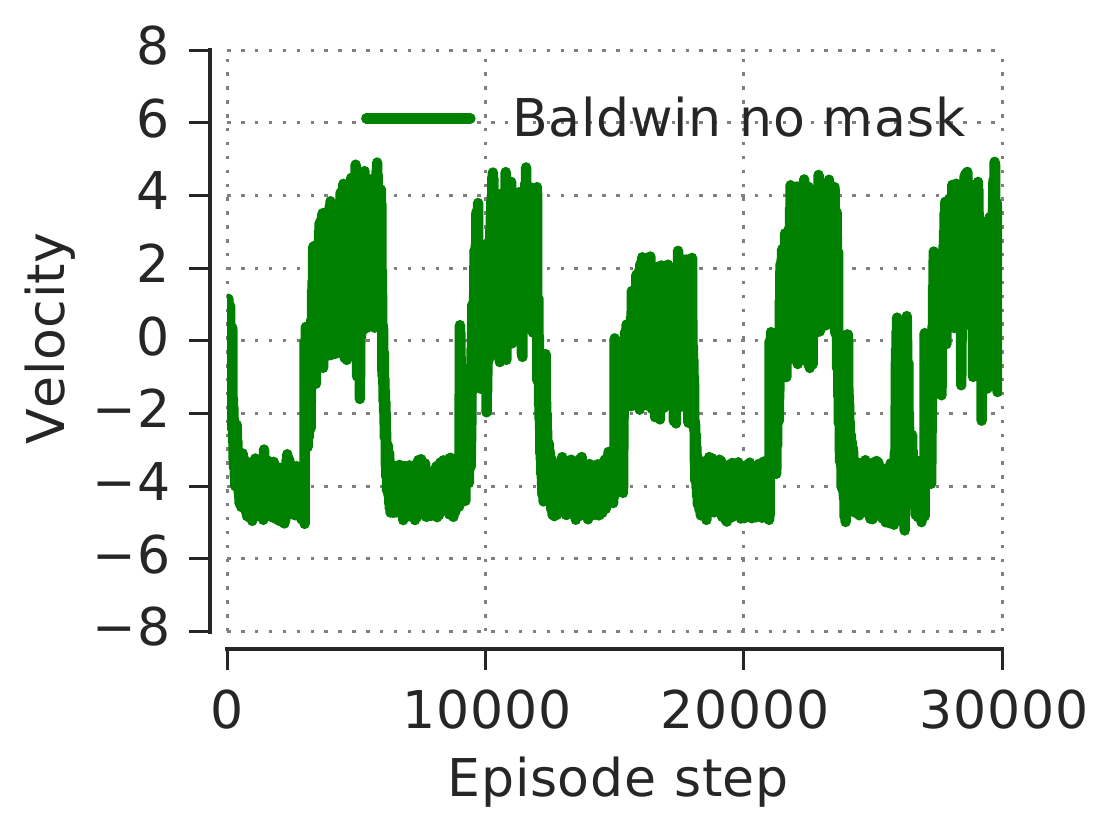}
    \caption{Baldwin, no mask}
  \end{subfigure}
  \begin{subfigure}{0.19\textwidth}
%    }
    \includegraphics[width=0.9\textwidth,valign=c]{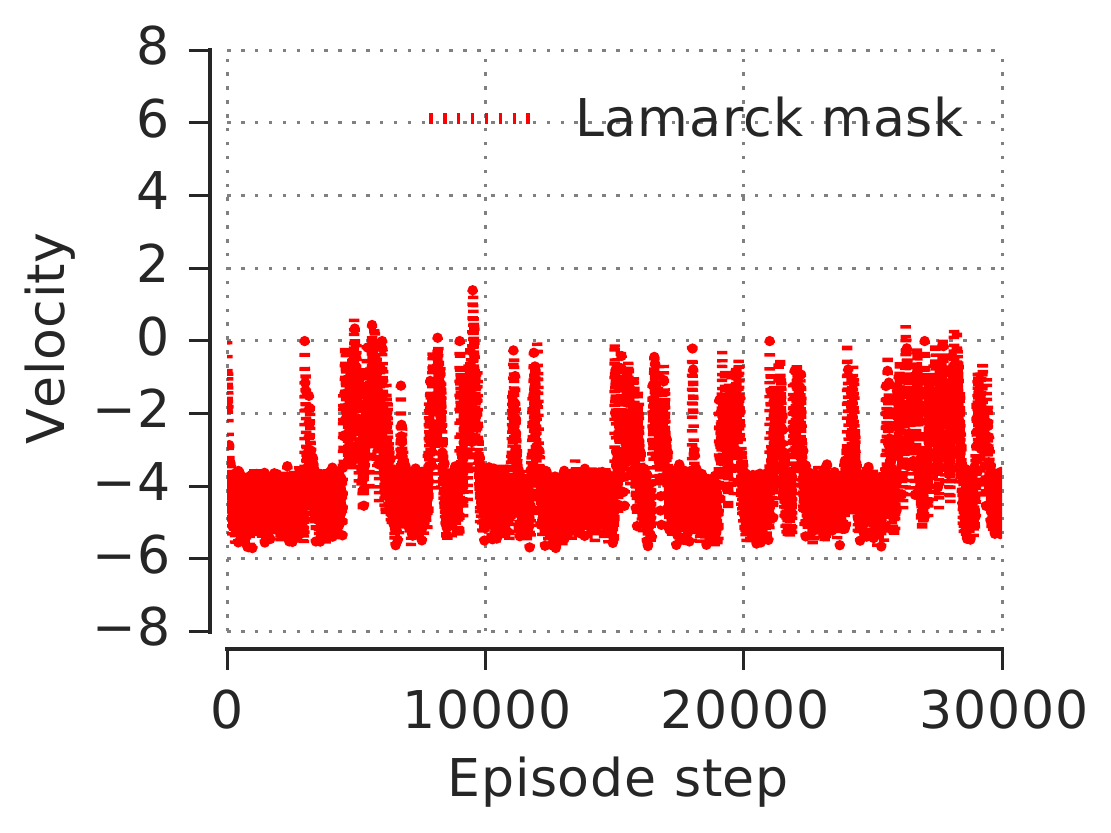}
    \caption{Lamarck, mask}
  \end{subfigure}
  \begin{subfigure}{0.20\textwidth}
%    }
    \includegraphics[width=0.9\textwidth,valign=c]{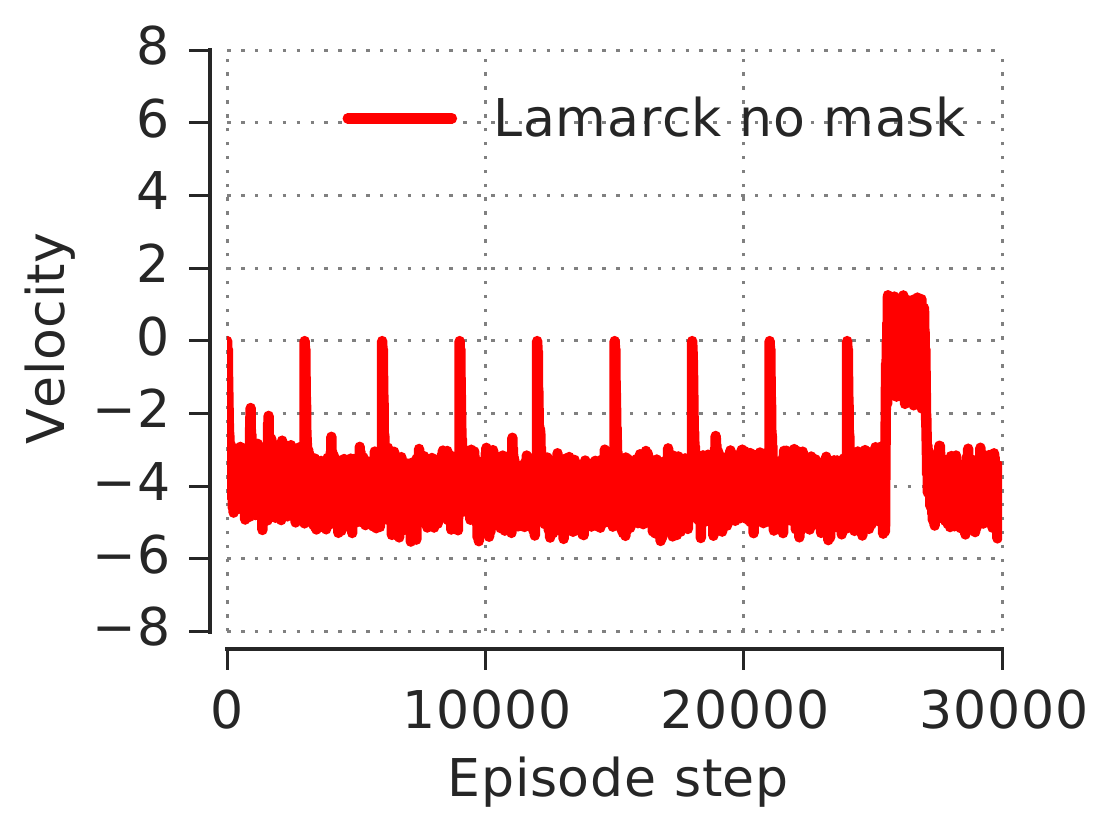}
    \caption{Lamarck, no mask}
  \end{subfigure}
  \begin{subfigure}{0.19\textwidth}
%    }
    \includegraphics[width=0.9\textwidth,valign=c]{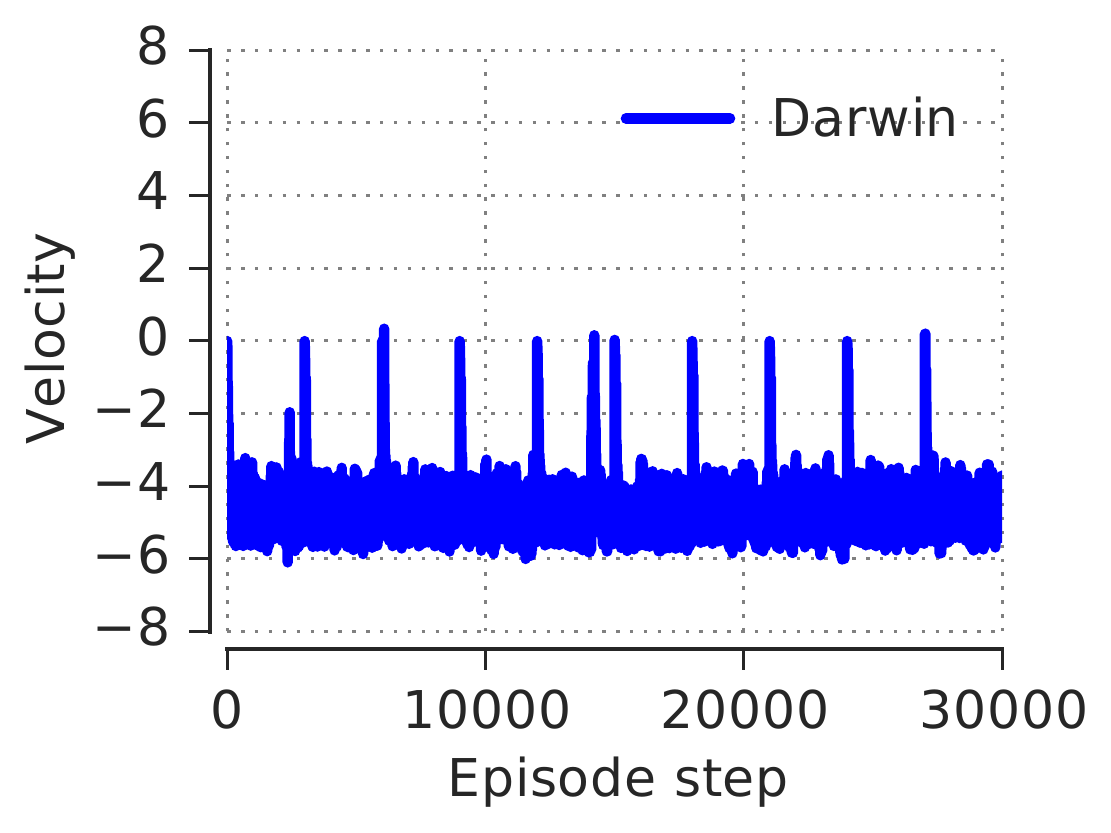}
    \caption{Darwin}
  \end{subfigure}
 %   }
    \caption{Velocities obtained in the Cheetah Goal Direction task for (a,b) Baldwinian, (c,d) Lamarckian, and (e) Darwinian evolution. Each fitness evaluation consists of 10 episodes with alternating requirements for backwards or forwards velocity. In the Baldwinian case, parameter values are reset at the start of each episode. In the Lamarckian case there is no resetting. The Baldwinian agents are capable of learning to go forwards and backwards as desired, but the Lamarckian agents evolve/learn only to go backwards. They have got stuck on that local optimum in this task. Video2Suppl shows the Baldwinian agent running forwards and backwards.}
    \label{GoalDirectionPerformance}
\end{figure*}

Figure \ref{sine-accomodation} shows the rate at which the neural network fits a particular sine wave presented during a lifetime. Similar to MAML, our methods's adaptation speed is superior to the one of the baseline approach (pretrained), which was trained to predict sine waves using a standard supervised-learning approach.\footnote{Plots for MAML and the pretrained approach come from \cite{finn2017model}.}

\subsection{Reinforcement Learning in Physics Environments}

\noindent {\bf Goal Velocity Task.} The Baldwin effect evolves a model that can quickly adapt its velocity to the target velocity within a single episode lasting only 30 simulated seconds. Figure \ref{GoalVelocity} shows that Lamarckian evolution outperforms Baldwinian evolution, which in turn outperforms Darwinian evolution. Figure \ref{GoalVelocityProfile} shows that Lamarckian evolution achieves the target velocity in each episode better than Baldwinian evolution. Both the Baldwin effect and Lamarckian learning are superior to pure Darwinian learning in this case.\\

\noindent {\bf Goal Direction Task.} The Baldwin effect evolves a model that can quickly adapt its direction to the target direction within a single episode lasting only 30 simulated seconds. Figure \ref{GoalDirectionSummary} shows the best agent fitness recorded over five independent evolutionary runs; two that use Baldwinian evolution (green), two that use Lamarckian evolution (red) and one that uses Darwinian evolution (blue), on the goal direction task. Best performance is obtained by Baldwinian evolution without an explicit plasticity mask, and second best with Baldwinian evolution with an explicit plasticity mask, followed by Darwinian evolution, with Lamarckian evolution a very clear loser in this task. The horizontal velocity of the cheetah over the course of one fitness evaluation is shown in Figure \ref{GoalDirectionPerformance}.

The contrast between the goal velocity and goal direction tasks is interesting. The goal direction task requires a radical change in policy for moving forwards or backwards in different episodes. Lamarckian evolution gets stuck in a local optimum of only being able to go backwards. Baldwinian evolution is able to cope with these two diverse tasks. In the goal velocity task, Lamarckian evolution is superior because the final velocity achieved in task $T-1$ is a suitable starting point for the target velocity required in task $T$ (note we increment the target velocity by $0.2$ in each episode). 

\begin{figure}[h]
  \centering
\includegraphics[width=3.3in,keepaspectratio=True]{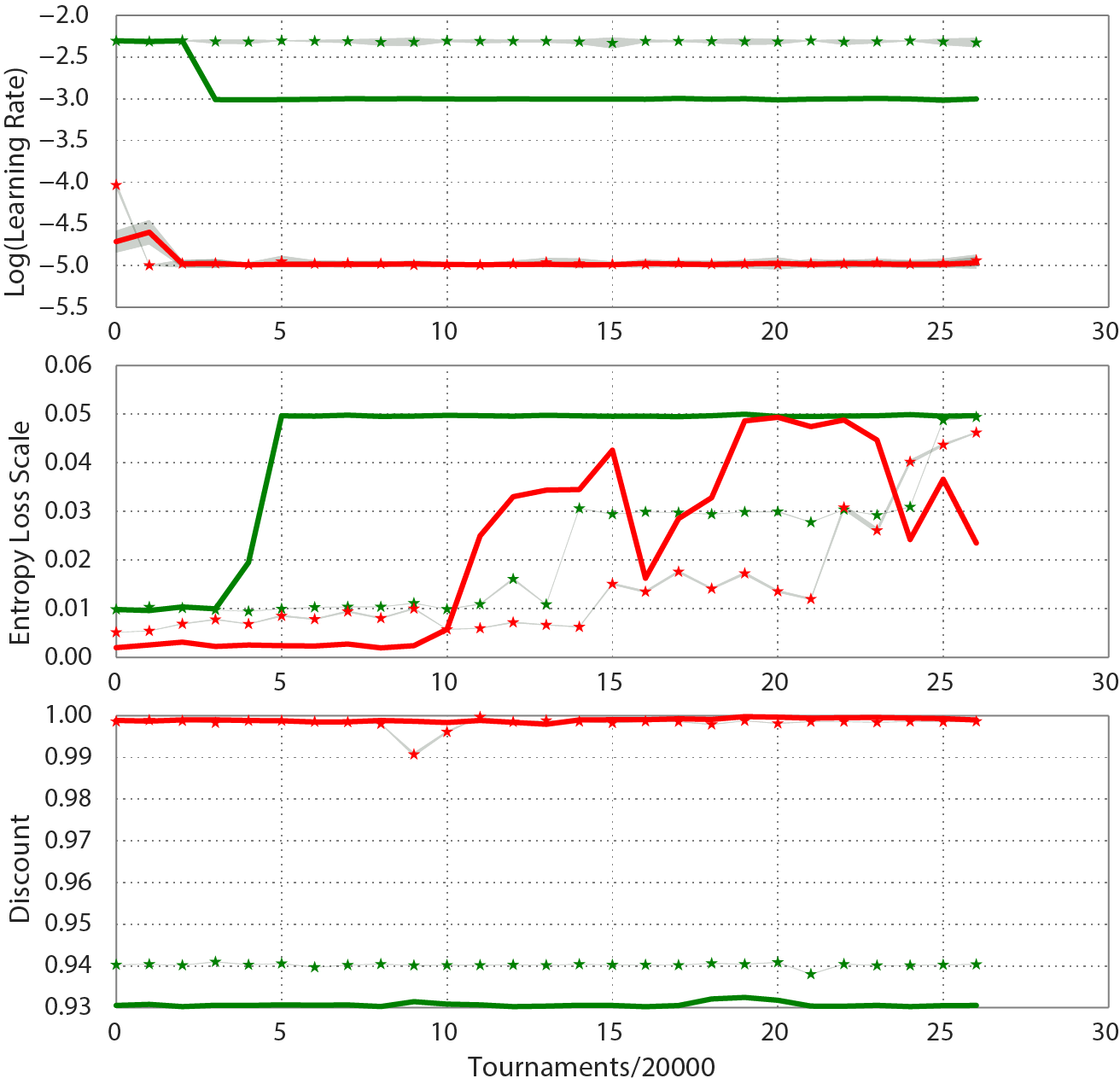}
\caption{Hyperparameter evolution shown at 25 evenly spaced timepoints in each evolutionary run. Green lines denote Baldwinian training, red lines denote Lamarckian training, averaged over a population of 500. Stars show with mask, and solid shows without mask. High learning rates evolve with Baldwin, but low learning rates evolve with Lamarck. Low discounts evolve with Baldwin, but high discounts evolve with Lamarck. There is no selection for entropy loss scale when a mask is used.}
\label{hyper1}
\end{figure}

How do the hyperparameters of the a2c algorithm evolve during the goal direction task? Figure \ref{hyper1} shows histograms of the distribution of hyperparameters for $5$ evenly spaced time-points during the runs. The main points to note are that in Baldwinian evolution we see learning rates evolve to quite high values, e.g. $0.005$ to $0.01$, whereas in Lamarckian evolution we see learning rates drop to the lowest values i.e. $0.00001$. In Baldwinian evolution the entropy loss scale evolves to high values $0.1$, but experiences little directed selection in Lamarckian evolution. In Baldwinian evolution the discounts become as small as we allow, i.e. 0.92, but in Lamarckian evolution they become as large as we allow i.e. $0.9999$. The Baldwin effect does not abolish learning in this task -- instead, it increases the rate of learning, but evolves strong learning biases. This is something that would not be possible in the Hinton and Nowlan task.

\section{Discussion and Conclusion}

In conclusion, in supervised learning tasks there was genetic accommodation of the initial function prior to learning, i.e. the regression network's prior was initially sinusoidal. Rapid learning continued to be selected for throughout evolution. In learning RL task distributions using the Baldwin effect, we observed that learning hyperparameters also evolved high learning rates and low discount factors, with the initial behaviour `at birth' providing strong biases to the learning algorithm which continued to show rapid learning throughout evolution. There is no complete genetic accommodation because that can never achieve high fitness by construction. Instead, it is the biases of the learning algorithm which are accommodated. The Baldwin effect is superior to Lamarckian learning when the distribution of tasks is broad or quickly changing, whereas Lamarckian learning is superior when the task distribution is narrow. The use of an explicit Hinton and Nowlan type mask did not speed up learning or final performance in task distributions. 

We have demonstrated that the Baldwin effect is capable of producing learning algorithms and models capable of few shot learning when combined with deep learning in supervised and reinforcement learning tasks. Further work is to show that this principle can be used to achieve state of the art results in machine learning on more complex task distributions.

Remarkably, meta-learning through evolution enables the use of non-differentiable fitness functions, in contrast to popular meta-learning approaches. For example, the fitness function can be defined on different, potentially multi-modal data distributions, making it a prime candidate for multi-objective optimization, even when data from one or several objectives is not always available to the low level optimization process.

%% file: algos.tex
\begin{algorithm}[t]
\caption{Model-Agnostic Meta-Learning (from \cite{finn2017model}) }
\label{alg:maml}
\begin{algorithmic}[1]
\REQUIRE $p(\task)$: distribution over tasks
\REQUIRE $\alpha$, $\beta$: step size hyperparameters
\STATE randomly initialize parameters $\theta$
\WHILE{not done}
\STATE Sample batch of tasks $\task_i \sim p(\task)$
  \FORALL{$\task_i$}
 \STATE Evaluate $\nabla_\theta \lossi(\theta)$ with respect to $K$ examples
 \STATE Compute adapted parameters with gradient descent:\\ $\theta_i'=\theta-\alpha \nabla_\theta  \lossi(\theta )$
 \ENDFOR
 \STATE Update $\theta \leftarrow \theta - \beta \, \nabla_\theta \sum_{\task_i \sim p(\task)}  \lossi ( {\theta_i'})$
\ENDWHILE
%\STATE while 
\end{algorithmic}
\end{algorithm}

\begin{algorithm}[t]
\caption{Baldwinian Meta-Learning}
\label{alg:baldwin_maml}
\begin{algorithmic}[1]
\REQUIRE $p(\task)$: distribution over tasks
\REQUIRE $\population$: initial population-representation of individuals
\REQUIRE $\getpopbatch$: procedure to obtain a batch of individuals (i.e. parameters and hyper-parameters) given a population-representation
\REQUIRE $\popupdate$: procedure to update a population-representation given a batch of fitness-scored individuals
\REQUIRE $\fitness$: fitness scoring function
\REQUIRE $N$: number of gradient steps to take during per-task gradient training
\WHILE{not done}
\STATE Generate batch of individuals  from population: \newline ${\theta^{g,\emptyset}}, \alpha^g \sim \getpopbatch(\population)$
\FORALL{$\theta^{g,\emptyset}$}
\STATE Sample batch of tasks $\task_i \sim p(\task)$
  \FORALL{$\task_i$}
  \STATE ${\theta^{g,i}} \leftarrow {\theta^{g,\emptyset}}$
  \FOR{$\texttt{k=1...N}$}
    \STATE Evaluate $\nabla_{\theta^{g,i}} \lossi({\theta^{g,i}})$ 
    \STATE Update adapted parameters with gradient descent:\newline $\theta^{g,i}\leftarrow \theta^{g,i}-\alpha^g \nabla_{\theta^{g,i}} \lossi({\theta^g,i} )$
  \ENDFOR
 \STATE Compute fitness-score for current task: 
 $f_i^g = \fitness(\theta^{g,i})$
 \ENDFOR
 \STATE Compute overall fitness estimate:
 $f^g = \sum_i f_i^g$
 \ENDFOR
 \STATE Update population based on fitness of individuals:\newline
 $\population \leftarrow \popupdate(\population,\{\left(1,{\theta^1},\alpha^1,f^1\right),..., \left(g,{\theta^g}, \alpha^g,f^g\right)\}$)
\ENDWHILE
%\STATE while 
\end{algorithmic}
\end{algorithm}